\documentclass[lettersize,journal]{IEEEtran}
\usepackage[table,dvipsnames]{xcolor}
\usepackage{amsmath,amsfonts}
\usepackage{algorithmic}
\usepackage{algorithm}
\usepackage{array}
\usepackage[caption=false,font=normalsize,labelfont=sf,textfont=sf]{subfig}
\usepackage{textcomp}
\usepackage{stfloats}
\usepackage{url}
\usepackage{verbatim}
\usepackage{graphicx}
\usepackage{cite}
\usepackage{multirow}
\usepackage{booktabs}

\definecolor{tablehighlight}{RGB}{242,243,243} 

\begin{document}

\title{Point Clouds Are Specialized Images: A Knowledge Transfer Approach for 3D Understanding}

\author{Jiachen Kang, Wenjing Jia, Xiangjian He, Kin Man Lam
\thanks{Corresponding author: Wenjing Jia.}
\thanks{J. Kang and W. Jia are with School of Electrical and Data Engineering, University of Technology Sydney, Australia. E-mail: Jiachen.Kang@student.uts.edu.au, Wenjing.Jia@uts.edu.au}
\thanks{X. He is with School of Computer Science, University of Nottingham Ningbo, China. E-mail: Sean.He@nottingham.edu.cn}
\thanks{K. M. Lam is with Department of Electrical and Electronic Engineering, The Hong Kong Polytechnic University, Hong Kong, China. E-mail: enkmlam@polyu.edu.hk}
}

\maketitle

\begin{abstract}
    Self-supervised representation learning (SSRL) has gained increasing attention in point cloud understanding, in addressing the challenges posed by 3D data scarcity and high annotation costs.
    This paper presents PCExpert, a novel SSRL approach that reinterprets point clouds as ``specialized images''.
    This conceptual shift allows PCExpert to leverage knowledge derived from large-scale image modality in a more direct and deeper manner, via extensively sharing the parameters with a pre-trained image encoder in a multi-way Transformer architecture. 
    The parameter sharing strategy, combined with a novel pretext task for pre-training, \textit{i.e.}, transformation estimation, empowers PCExpert to outperform the state of the arts 
    in a variety of tasks, with a remarkable reduction in the number of trainable parameters.
    Notably, PCExpert's performance under \texttt{LINEAR} fine-tuning (\textit{e.g.}, yielding a 90.02\% overall accuracy on ScanObjectNN) has already approached the results obtained with \texttt{FULL} model fine-tuning (92.66\%), demonstrating its effective 
    and robust representation capability. 
\end{abstract}

\section{Introduction}\label{s:intro}

Point cloud, a data format that uses coordinates and other attributes to represent three-dimensional objects, has demonstrated significant potential in deep learning and found wide-ranging applications.
However, the acquisition of point cloud data is still inconvenient, because scanning equipment's design is usually aimed toward professional needs, and the scanning process is more complex than 2D photo capturing~\cite{raj2020survey}.
Furthermore, annotating the labels (ground truth) of 3D data for supervised learning tasks is typically more complex and time-consuming than 2D image data~\cite{behley2019semantickitti}. 
As a result, point cloud datasets tend to be smaller in terms of the number of individual samples, and only using annotated data may not be sufficient for point cloud understanding and applications. 
In order to better comprehend point cloud data while circumventing time-consuming data annotation, point cloud self-supervised representation learning (SSRL) has garnered growing attention in recent years. 
This paradigm sidesteps the need for data annotation and, with properly designed models and pretext tasks, can yield performance comparable to supervised approaches.

Currently, SSRL encompasses two popular approaches: contrastive-based~\cite{zhang2021self,afham2022crosspoint,sun2023vipformer} and reconstruction-based~\cite{wang2021unsupervised,yu2022point,pang2022masked,dong2023autoencoders}.
Since reconstruction-based approaches do not require positive or negative samples, they are more feasible to implement 
and thus have recently received prominence in point cloud understanding studies.
However, with the current substantial advancements in multi-modal learning~\cite{radford2021learning,xu2022semantic,wang2023image,girdhar2023imagebind}, we identify novel opportunities to enhance SSRL's effectiveness using contrastive objectives.
Various studies~\cite{radford2021learning,bao2022vlmo} have shown that the representational capacity of image models can be significantly enhanced when they are aligned with large volumes of textual data.
The alignment has even led to impressive performance in zero-shot classification scenarios.
In these explorations, image data is studied as if it were a ``foreign language''~\cite{wang2023image}. This naturally provokes a question: can 
point clouds 
be regarded as specialized images?
Motivated by this question, the present study pursues a point-image contrastive-based approach to point cloud understanding.

The standpoint of considering point cloud data as ``specialized images'' brings a paradigm shift in our mindset towards the design of architectures.
Firstly, to address the aforementioned issue of the scarcity of point cloud datasets, we propose that models pre-trained on large-scale image datasets, instead of point datasets, can also serve a crucial role in guiding point cloud learning.
This proposition is supported by recent research~\cite{gao2023ulip,qi2023recon}, where the CLIP model~\cite{radford2021learning} was employed as guidance.
Secondly, in order to transfer knowledge more effectively from image modality to point cloud modality, we assume that a substantial degree of parameter sharing between modalities is helpful. 
Previous studies on point-image contrastive learning generally utilize separate encoders for each modality independently~\cite{jing2021cross,li2022closer,afham2022crosspoint,li2022simipu,gao2023ulip,qi2023recon}. 
This facilitates the adaptation of inductive biases to the unique characteristics of each modality.
However, these methods miss the potential to apply knowledge acquired from large-scale images to point data at a deeper level with parameter sharing.

\begin{figure*}[!t] 
    \vskip 0.1in
    \begin{center}
    \centerline{\includegraphics[width=1\textwidth]{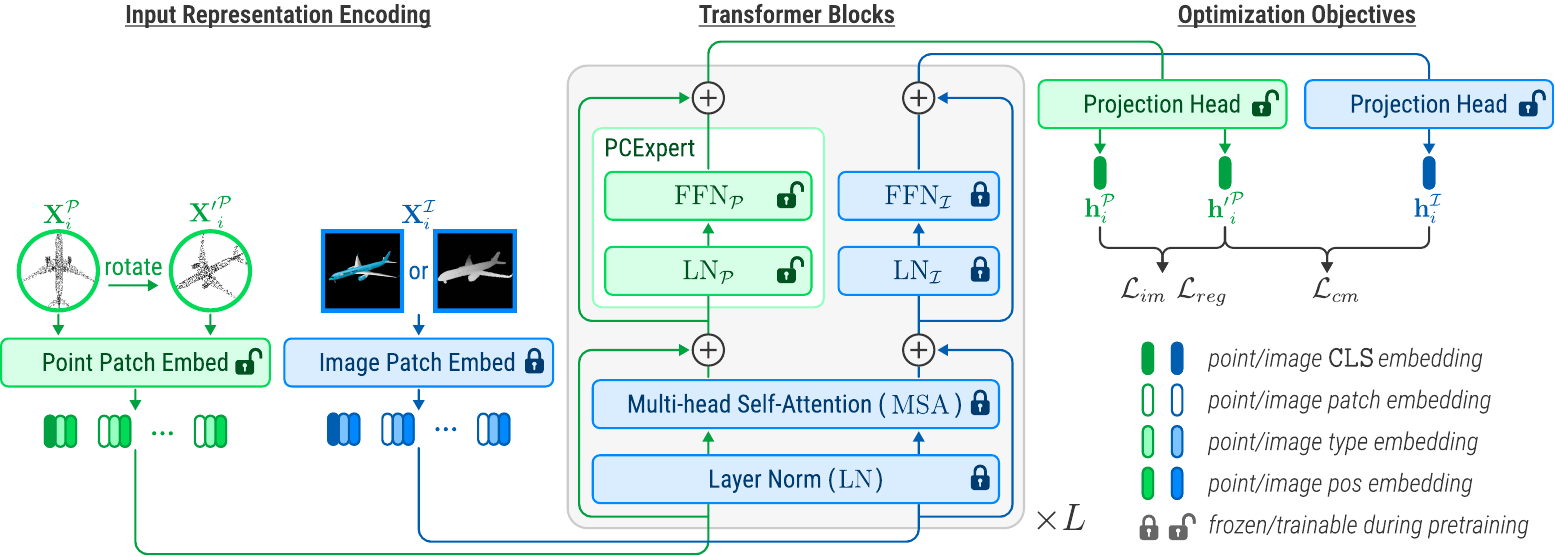}} 
    \caption{The pipeline of PCExpert. 
    \textbf{Left}: The input representations consist of sequences of embeddings, which are the summation of the patch/\texttt{CLS} tokens, the type embeddings and the position embeddings for the respective point and image data.
    \textbf{Middle}: The point and image input representations are then fed into a series of transformer blocks. 
    In each block, the representations are first processed by a shared Vision Transformer (ViT) Multi-head Self-Attention (MSA) module, and then processed by separate Feed Forward Networks (FFNs), according to their modality.
    \textbf{Right}: During the pre-training process, the parameters in ViT are kept frozen, while only the parameters related to point processing and projection heads are optimized, via three objectives: cross-modal contrastive ($\mathcal{L}_{cm}$), intra-modal contrastive ($\mathcal{L}_{im}$) and rotation angle regression ($\mathcal{L}_{reg}$).
    }
    \label{f:architecture}
    \end{center}
    \vskip -0.3in
\end{figure*}

In this study, the multi-way Transformer~\cite{bao2022vlmo} is adopted for point-image contrastive learning.
Throughout the encoding of image and point data, this architecture enables an extensive sharing of parameters belonging to the image encoder, while providing a modular network for the acquisition of point cloud-specific knowledge.
As this modular network is solely dedicated to the processing of point cloud data, we call it ``PCExpert'' (Point Cloud Expert) in this study. 
Fig.~\ref{f:architecture} illustrates the pipeline of our proposed PCExpert architecture, detailing three key components: 1) the process of input representations, 2) the integration of PCExpert within Transformer blocks, and 3) the employed learning objectives for SSRL. 
Furthermore, PCExpert can also be conceptualized as a plug-in system for pre-trained Transformers. 
This system extends the network's functionality to a new modality with only a marginal increment in the number of parameters, while preserving the performance of the original model.

In addition to the proposed PCExpert architecture, this study also introduces a novel pretext task for SSRL.
Specifically, drawing from the insights in a prior study that learning factors of variation can concurrently enhance invariance learning~\cite{kang4200653learning}, we hypothesize that the task of estimating transformation parameters can be a good complement to the conventional contrastive learning objectives.
Therefore, we propose to minimize  ``regression loss'' during the estimation of the transformation parameters, and enforce the learning of more descriptive representations which are capable of differentiating point clouds by leveraging their intrinsic relationships.  

In experimental results, PCExpert exhibits robust representational capacity, with a much lower number of parameters in comparison to the current SSRL methods. 
Combined with the regression loss, the model has achieved state-of-the-art (SOTA) results across several 
benchmarks. 
For instance, in the real-world dataset ScanObjectNN, PCExpert achieved an overall accuracy (OA) of $90.02\%$ in the \texttt{LINEAR} fine-tuning protocol, with a $5\%$ improvement over the 
SOTA performance.
Furthermore, we have also taken into account the circumstance where the dataset lacks contrastive images.
We conducted a parallel series of pre-training that is solely based on the 3D modality, using images rendered directly from point clouds.
Even in this scenario, PCExpert is still capable of achieving performance on par with the established benchmarks.

Our main contributions in this study are as follows:
\begin{itemize}
    \item We propose PCExpert for point cloud SSRL.
    To the best of our knowledge, it is the first architecture that exploits both image knowledge guidance and extensive parameter sharing in image-point contrastive learning. 
    PCExpert provides evidence that Transformer blocks for image encoding are also capable of directly encoding point clouds, 
    thus allowing knowledge of large-scale image data to be utilized for point cloud understanding, in a more intricate manner. 
    \item We introduce an alternative approach for creating datasets for image-point contrastive learning, which reduces the cost and difficulty associated with data collection. 
    Our research indicates that for the positive sample pair, mesh-rendered images are not essential. 
    Instead, images directly rendered by point clouds can be used as positive samples, with only a minimal impact on performance.
    \item In the pre-training phase, we introduce 
    transformation parameter estimation as a new, extra pretext task, leveraging on the regression loss. 
    Alongside the contrastive objectives, this task further enhances model performance.
\end{itemize}

In summary, our research demonstrates 
that point cloud understanding can be reconceptualized and realized 
as the understanding of ``specialized images''. 
More importantly, the substantial advancements in current multi-modal learning are significantly driven by 1) the exploitation of large-scale datasets and 2) the scalability and versatility of Transformers.
With this perspective, our work presents a promising pathway towards more effective self-supervised point cloud understanding through image-assisted cross-modal learning, leveraging the potential of large-scale, low-cost datasets and pre-trained multi-modal Transformers. 

The remainder of this paper is organized as follows. 
Section~\ref{s:related} provides a comprehensive review of related work, focusing on multi-modal studies on point cloud learning. 
Section~\ref{s:methods} presents our proposed PCExpert, outlining the key components and algorithms. 
In Section~\ref{s:experiment}, we illustrate the experimental setup and present the evaluation results, as well as ablation studies, and discussions on our findings. 
Finally, Section~\ref{s:conclusion} concludes the paper, summarizing the key findings and discussing future directions. 

\section{Related Works}\label{s:related}

\subsection{Contrastive Learning for Point-image Modality} 
\noindent A major subset of point cloud SSRL methodologies are based on contrastive learning principles.
The primary objective of these approaches is to maximize the agreement between different views of the same 3D object while simultaneously minimizing the agreement between unrelated ones.
An effective ingredient to this learning paradigm is harnessing the image modality to provide complementary information for point cloud understanding~\cite{liu2021learning,jing2021cross,li2022closer,afham2022crosspoint,li2022simipu,zhou2022pointcmc,wu2023self}.
For instance, Jing \textit{et al.}~\cite{jing2021cross} introduced the Center loss, aimed at aligning features across multiple modalities.
Some studies~\cite{hess2022lidarclip,zhang2023clip} leveraged a pre-existing embedding distribution with a pre-trained image model, to guide point cloud feature distillation.

A notable contribution by Afham \textit{et al.}~\cite{afham2022crosspoint} was the proposition of  intra- and cross-modal contrastive loss, which enhances point-image alignment and point instance discrimination, simultaneously.
As distinct from the instance-level contrastive, Li \textit{et al.}~\cite{li2022simipu} proposed a patch-level contrastive approach for better spatial comprehension, using the Hungarian Algorithm.
Moreover, some research works~\cite{liu2021learning,li2022closer} advocate pixel/point-level contrastive learning to facilitate local feature correspondence.
Zhou \textit{et al.}~\cite{zhou2022pointcmc} proposed multi-scale contrastive objectives between multi-modality objects, enabling local-to-global feature alignment.

Our study adopts the pretext tasks from~\cite{afham2022crosspoint} and builds upon them.
Additionally, we introduce a unique task that leverages a ``regression loss'' for transformation parameter estimation.
To the best of our knowledge, we are the first to apply this objective to point cloud representation learning.

\subsection{Knowledge Transfer with Pre-trained Image Models} 
\noindent The above-mentioned methodologies predominantly rely on feature alignment. 
As a consequence, it is intuitive to utilize separate feature extractors for each modality. 
Instead of exploiting feature-level guidance for knowledge transfer, an alternate strand of research seeks to directly conduct point cloud understanding with pre-trained \textit{image} models.

For instance, Xu \textit{et al.}~\cite{xu2022image2point} proposed an ``inflating'' method to convert 2D convolutional networks, pre-trained on image datasets, to 3D convolutional networks, thus catering to point cloud/voxel processing.
Studies in~\cite{zhang2022pointclip,wang2022p2p} conducted point cloud analysis, by transforming point data into images that are recognizable by pre-trained image models.
To realize the transformation, techniques including geometry-preserving projection with geometry-aware coloring~\cite{wang2022p2p}, and multi-view projection~\cite{zhang2022pointclip} are applied, respectively.
Rong \textit{et al.}~\cite{rong2023efficient} utilized a pre-trained image semantic segmentation model to process images rendered from point clouds in 3D scene segmentation tasks.
Other studies~\cite{qian2022pix4point,huang2022frozen} directly feed tokenized point data into a pre-trained image model for supervised learning.
Moreover, Dong \textit{et al.}~\cite{dong2023autoencoders} leveraged a pre-trained image model as a cross-modal teacher during point cloud masked modelling.

In contrast to the aforementioned methodologies, our approach neither employs separate encoders for each modality, nor uses the same model for both modalities.
Instead, based on our perspective that point clouds are ``specialized images'', 
extensive parameter sharing is realized between the point and image encoders, with additional sets of trainable parameters provided for point-specific knowledge acquisition.

\section{The Proposed Method}\label{s:methods}

\noindent As depicted in Fig.~\ref{f:architecture}, 
the architecture of the PCExpert module, combined with the pre-trained Vision Transformer (ViT), serves as a foundation for processing multi-modal inputs, \textit{i.e.}, point and image data. 
Before being fed into the transformer blocks, point and image data are initially embedded in a D-dimensional space as sequential input representations.
Within each transformer block, the point and image input representations are first processed by 
the Multi-head Self-Attention (MSA) module of the original ViT. 
The representations are then subjected to parallel projection paths in separate feed forward networks (FFNs), according to their modality. 
During point cloud SSRL, the parameters in ViT are kept frozen, while only the parameters of PCExpert and the projection heads are optimized, via three objectives: cross-modal contrastive, intra-modal contrastive and transformation parameter (\textit{i.e.}, the rotation angle) regression.
This architecture and training strategy focus on point-specific representation learning with extensive image-to-point knowledge transfer, without affecting the original ViT performance on image-related tasks. 
This also significantly reduces computation and storage requirements, as only a small fraction ($\approx6.6\%$) of parameters are updated.

In the following sections, we provide a concrete explanation of the construction of input representations (Section~\ref{s:methods:input}), the architecture of PCExpert (Section~\ref{s:methods:pcexpert}) and the pre-training process of point cloud SSRL (Section~\ref{s:methods:training}).

\subsection{Input Representations}
\label{s:methods:input}

\noindent During point cloud SSRL, a training batch comprises $N$ triplets, \textit{i.e.}, 
$\{\mathbf{X}^\mathcal{P}_i, \mathbf{X'}^\mathcal{P}_i, \mathbf{X}^\mathcal{I}_i\}^N_{i=1}$, where the superscripts $\mathcal{P}$ and $\mathcal{I}$ denote the point and image modalities\footnote{This notation is maintained consistently throughout this paper to signify the two modalities.}, respectively.
To generate data for the intra-modal contrastive objective (see Section~\ref{s:methods:training:im_loss}), each point data $\mathbf{X}_\mathcal{P}$ is applied with a random transformation, resulting in $\mathbf{X'}_\mathcal{P}$.
Then, point and image data are embedded to derive sequential input representations, as the input for the Transformer blocks.
We now explain the embedding process. 

\subsubsection{Point Input Representations}\label{s:methods:input:point}
A point cloud $\mathbf{X}_\mathcal{P} \in \mathbb{R}^{M \times 3}$ 
(or $\mathbf{X'}_\mathcal{P}$)
consists of $M$ points defined by coordinates in an $(x,y,z)$ Cartesian space.
Following the previous study in~\cite{yu2022point}, we sample $N_\mathcal{P}$ centroids using farthest point sampling (FPS).
To each of these centroids, we assign $k$ neighbouring points by conducting a $k$-nearest neighbour (kNN) search.
Thereby, we obtain $N_\mathcal{P}$ local geometric groups 
$\{G_i\}^{N_\mathcal{P}}_{i=1}$, where each group $G_i$ consists of a centroid
$\mathbf{x}^{\mathcal{P}}_{i,0}$, and its $k$ neighboring points
$\{\mathbf{x}^{\mathcal{P}}_{i,j}\}^k_{j=1}$, \textit{i.e.},
$G_i = \{\mathbf{x}^{\mathcal{P}}_{i,j}\}^k_{j=0}$.

The patch embeddings 
$\{\mathbf{Z}^{\mathcal{P}}_i\}^{N_\mathcal{P}}_{i=1}$ 
for $\{G_i\}^{N_\mathcal{P}}_{i=1}$
are extracted with a two-layer PointNet++~\cite{qi2017pointnet++},
where $\mathbf{Z}^{\mathcal{P}}_i \in \mathbb{R}^D$ and D is the embedding size.
Concretely, for $j = 1, ..., N_\mathcal{P}$, 
\begin{subequations}
\begin{align}
    \Tilde{\mathbf{Z}}^{\mathcal{P}}_i & = \max_{\mathbf{x}^{\mathcal{P}}_{i,j} \in G_i} [ \, f_1(\mathbf{x}^{\mathcal{P}}_{i,j} \, ; \, \mathbf{x}^{\mathcal{P}}_{i,j} - \mathbf{x}^{\mathcal{P}}_{i,0}) \, ]\\
    \mathbf{Z}^{\mathcal{P}}_i & = \max_{\mathbf{x}^{\mathcal{P}}_{i,j} \in G_i} [ \, f_2(\mathbf{x}^{\mathcal{P}}_{i,j} \, ; \, \Tilde{\mathbf{Z}}^{\mathcal{P}}_i) \, ],
\end{align}
\end{subequations}
where $f_1$ and $f_2$ are Multi-layer Perceptrons (MLPs).
A learnable class embedding 
$\mathbf{Z}^{\mathcal{P}}_{\mathtt{CLS}} \in \mathbb{R}^D$ 
is prepended to the sequence of the patch embeddings.

To obtain the point input representations
$\mathbf{H}^{\mathcal{P}}_0 \in \mathbb{R}^{(N_\mathcal{P} + 1) \times D}$,
we sum the sequence of patch embeddings with point position embeddings 
$\mathbf{Z}^{\mathcal{P}}_{pos} \in \mathbb{R}^{(N_\mathcal{P} + 1) \times D}$
and a point type embedding
$\mathbf{Z}^{\mathcal{P}}_{type} \in \mathbb{R}^D$:
\begin{equation}\label{e:pcd_input}
    \mathbf{H}^{\mathcal{P}}_0 = [\mathbf{Z}^{\mathcal{P}}_{\mathtt{CLS}}, \: \mathbf{Z}^{\mathcal{P}}_1, \: ..., \: \mathbf{Z}^{\mathcal{P}}_{N_\mathcal{P}}] + \mathbf{Z}^{\mathcal{P}}_{pos} + \mathbf{Z}^{\mathcal{P}}_{type}
\end{equation}
Position embeddings 
$\mathbf{Z}^{\mathcal{P}}_{pos}$
are derived by applying a two-layer MLP on centroid points 
$\{\mathbf{x}^{\mathcal{P}}_{i,0}\}^{N_\mathcal{P}}_{i=1}$.
In the case of
$\mathbf{Z}^{\mathcal{P}}_{\mathtt{CLS}}$,
a virtual centroid 
with coordinates set at $(0,0,0)$ is used to generate the positional embedding.

\subsubsection{Image Input Representations}\label{s:methods:input:image}
We follow the studies in~\cite{dosovitskiy2020image,bao2022vlmo} and split the image data
$\mathbf{X}_\mathcal{I} \in \mathbb{R}^{H \times W \times C}$
into $N_\mathcal{I}$ patches
$\{\mathbf{x}^{\mathcal{I}}_i\}^{N_\mathcal{I}}_{i=1}$, where
$N_\mathcal{I} = HW/P^2$, 
$\mathbf{x}^{\mathcal{I}}_i \in \mathbb{R}^{P^2 \times C}$, 
$C$ is the number of channels, and $(H,W)$ and $(P,P)$ are the resolutions of the image and patches, respectively.
The sequence of image patch embeddings
$\{\mathbf{Z}^{\mathcal{I}}_i\}^{N_\mathcal{I}}_{i=1}$
are linearly projected from these patches: 
$\mathbf{Z}^{\mathcal{I}}_i = \mathbf{V} \mathbf{x}^{\mathcal{I}}_i$
with $\mathbf{V} \in \mathbb{R}^{(P^2 \times C) \times D}$.

Similar to the point input representations, the image input representations 
$\mathbf{H}^{\mathcal{I}}_0 \in \mathbb{R}^{(N_\mathcal{I} + 1) \times D}$
are calculated by summing the image patch embeddings
(prepended by the class embedding $\mathbf{Z}^{\mathcal{I}}_{\mathtt{CLS}}$)
with image position embeddings 
$\mathbf{Z}^{\mathcal{I}}_{pos} \in \mathbb{R}^{(N_\mathcal{I} + 1) \times D}$
and an image type embedding
$\mathbf{Z}^{\mathcal{I}}_{type} \in \mathbb{R}^D$:
\begin{equation}\label{e:img_input}
    \mathbf{H}^{\mathcal{I}}_0 = [\mathbf{Z}^{\mathcal{I}}_{\mathtt{CLS}}, \: \mathbf{Z}^{\mathcal{I}}_1, \: ..., \: \mathbf{Z}^{\mathcal{I}}_{N_\mathcal{I}}] + \mathbf{Z}^{\mathcal{I}}_{pos} + \mathbf{Z}^{\mathcal{I}}_{type}
\end{equation}

\subsection{Point Cloud Expert}
\label{s:methods:pcexpert}

\noindent Inspired by previous works~\cite{fedus2022switch,bao2022vlmo}, we propose PCExpert (Point Cloud Expert), a specialized network for enhancing point cloud understanding by leveraging image knowledge.
We employ a pre-trained ViT to encode both point and image modalities.
Different from the standard ViT, our architecture incorporates separate FFNs, each dedicated to a specific modality (denoted by $\textrm{FFN}_\mathcal{P}$ and $\textrm{FFN}_\mathcal{I}$).
Concretely, if we denote by $\mathbf{H}^{\mathcal{P}}_{l-1}$ and $\mathbf{H}^{\mathcal{I}}_{l-1}$ the point and image input representations for the $l$-th transformer block, then the output representations for point cloud and image can be computed respectively as: 
\begin{subequations}
\begin{align}
\label{e:transformer1}
    \Tilde{\mathbf{H}}^{\mathcal{P}}_l &= \textrm{MSA} \, ( \, \textrm{LN} \, ( \, \mathbf{H}^{\mathcal{P}}_{l-1} \, )) + \mathbf{H}^{\mathcal{P}}_{l-1} \\ 
    \mathbf{H}^{\mathcal{P}}_l &= \textrm{FFN}_\mathcal{P} \, ( \, \textrm{LN}_\mathcal{P} \, ( \, \Tilde{\mathbf{H}}^{\mathcal{P}}_l \, )) + \Tilde{\mathbf{H}}^{\mathcal{P}}_l 
\end{align}
\end{subequations}
\begin{subequations}
\begin{align}
\label{e:transformer2}
    \Tilde{\mathbf{H}}^{\mathcal{I}}_l &= \textrm{MSA} \, ( \, \textrm{LN} \, ( \, \mathbf{H}^{\mathcal{I}}_{l-1} \, )) + \mathbf{H}^{\mathcal{I}}_{l-1} \\ 
    \mathbf{H}^{\mathcal{I}}_l &= \textrm{FFN}_\mathcal{I} \, ( \, \textrm{LN}_\mathcal{I} \, ( \, \Tilde{\mathbf{H}}^{\mathcal{I}}_l \, )) + \Tilde{\mathbf{H}}^{\mathcal{I}}_l
\end{align}
\end{subequations}
where $\textrm{LN}$ denotes the layer normalisation operation.
The mutual MSA module facilitates image knowledge sharing with the point modality, while the separate FFNs ensure that the unique features of each modality are effectively captured and integrated into the overall representations.

\subsection{Training}
\label{s:methods:training}

\noindent PCExpert is primarily trained with a point-image contrastive objective, in order to exploit the guidance offered by the image modality.
Furthermore, we follow the methodologies in~\cite{afham2022crosspoint} and implement an intra-modal contrastive learning, for the purpose of enhancing the learning of point semantic invariance.
Drawing inspiration from~\cite{kang4200653learning}, we integrate transformation parameter estimation as an additional pretext task for SSRL.
This enables the model to capture the causal knowledge embodied in the representations and to mitigate the influence of confounding factors of variation, thus refining the quality of the learned representations.

As for the optimisation process during training, only the parameters of PCExpert are updated via back-propagation, while the original parameters of ViT are frozen.
This training strategy ensures the focus of optimization of the model is dedicated to point data, and does not compromise the model performance on images, whilst offering significant benefits in terms of computational efficiency.

\subsubsection{Cross-modal Contrastive Learning}
\label{s:methods:training:cm_loss}
Given a batch comprising $N$ point-image pairs $\{\mathbf{X}^\mathcal{P}_i\}^N_{i=1}$ and $\{\mathbf{X}^\mathcal{I}_i\}^N_{i=1}$, the purpose of point-image contrastive learning is to discern the corresponding (positive) pairs from a pool of $N^2$ potential pairs.
The output [\texttt{CLS}] tokens of the final ($L$-th) transformer block
$\{\mathbf{H}^{\mathcal{P}}_{\mathtt{CLS},L,i}\}^N_{i=1}$ and 
$\{\mathbf{H}^{\mathcal{I}}_{\mathtt{CLS},L,i}\}^N_{i=1}$
are used as the global representations of the point and image data, respectively.

Subsequently, these [\texttt{CLS}] tokens are mapped to an invariant space via two projection heads $f_{\mathcal{P}}$ and $f_{\mathcal{I}}$, \textit{i.e.}, 
\begin{align}
    &\mathbf{h}^{\mathcal{P}}_i = f_{\mathcal{P}}(\mathbf{H}^{\mathcal{P}}_{\mathtt{CLS},L,i})\\
    &\mathbf{h}^{\mathcal{I}}_i = f_{\mathcal{I}}(\mathbf{H}^{\mathcal{I}}_{\mathtt{CLS},L,i})
\end{align}

The distances between the (normalized) output embeddings 
$\mathbf{h}^{\mathcal{P}}_i$ and $\mathbf{h}^{\mathcal{I}}_i$ 
in this space are calculated using cosine similarity.
Thereby, the loss function for the positive point-image pair is defined as:
\begin{equation}
    \mathcal{L}^{\mathcal{P}2\mathcal{I}}_{cm,i} = - \log \frac{\exp{({\mathbf{h}^{\mathcal{P}}_i} \, ^\mathsf{T} \, \mathbf{h}^{\mathcal{I}}_i/\tau)}}{ \sum^N_{j=1} \exp{({\mathbf{h}^{\mathcal{P}}_i} \, ^\mathsf{T} \, \mathbf{h}^{\mathcal{I}}_j/\tau)}},
\end{equation}
where $\tau$ stands for the temperature co-efficient, and the superscript $\mathcal{P}2\mathcal{I}$
signifies that, optimizing this loss facilitates the alignment of the $i$-th point with the corresponding image among $N$ images.
Similarly, if we denote by $\mathcal{I}2\mathcal{P}$ the reciprocal task to align an image with its corresponding point cloud,
the cross-modal contrastive loss $\mathcal{L}_{cm}$ is expressed as:
\begin{equation}
    \mathcal{L}_{cm} = \frac{1}{2N} \sum^N_{i=1} \, ( \, \mathcal{L}^{\mathcal{P}2\mathcal{I}}_{cm,i} + \mathcal{L}^{\mathcal{I}2\mathcal{P}}_{cm,i} \, )
\end{equation}

\subsubsection{Intra-modal Contrastive Learning}
\label{s:methods:training:im_loss}
In addition to aligning the features between point and image modality, we conduct contrast within the point cloud modality.

Given a batch of point cloud data $\{\mathbf{X}^\mathcal{P}_i\}^N_{i=1}$, 
we apply transformation $T$ on each sample to get $\{\mathbf{X'}^\mathcal{P}_i\}^N_{i=1}$.
A positive pair is defined as the original sample and its transformed version.
Similar to the point-image contrastive loss, the intra-modal contrastive loss $\mathcal{L}_{im}$ can be expressed as:
\begin{align}
    &\mathcal{L}^{\mathcal{P}2\mathcal{P'}}_{im,i} = - \log \frac{\exp{({\mathbf{h}^{\mathcal{P}}_i} \, ^\mathsf{T} \, \mathbf{h'}^{\mathcal{P}}_i/\tau)}}{ \sum^N_{j=1} \exp{({\mathbf{h}^{\mathcal{P}}_i} \, ^\mathsf{T} \, \mathbf{h'}^{\mathcal{P}}_j/\tau)}}\\
    &\mathcal{L}_{im} = \frac{1}{2N} \sum^N_{i=1} \, ( \, \mathcal{L}^{\mathcal{P}2\mathcal{P'}}_{im,i} + \mathcal{L}^{\mathcal{P'}2\mathcal{P}}_{im,i} \, )
\end{align}
where the superscripts $\mathcal{P}2\mathcal{P'}$ and $\mathcal{P'}2\mathcal{P}$ signify the original-to-transformed and transformed-to-original sample pair matching, respectively.

\subsubsection{Transformation Parameter Estimation}\label{s:methods:training:reg_loss}
Based on the findings of the study in~\cite{kang4200653learning}, the factors of variation in the data generation process have significant influence in causal inference.
With this understanding, we use the transformation $T$ in intra-modal contrastive learning as an additional supervisory signal to guide point understanding.
The objective of the transformation parameter estimation task is to perform numerical regression on the transformation $T$'s value.

Specifically, we apply $y$-axis rotation as the transformation in this study, thus making the rotation angle as the ground truth.
In order to circumvent numerical cycles caused by periodic symmetry in rotation, we quantize the rotation angles into $d$ categories, and project each category into a $\mathbb{R}^d$ space as a one-hot vector $\{y_i\}^N_{i=1}, y_i \in \mathbb{R}^d$. 

To calculate the regression loss, we first calculate the difference between 
$\mathbf{h}^{\mathcal{P}}_i$ and $\mathbf{h'}^{\mathcal{P}}_i$, 
and linearly project the resultant difference vector into the same $\mathbb{R}^d$ space using $f_T$, \textit{i.e.}, 
$\hat{y_i} = f_T(\mathbf{h}^{\mathcal{P}}_i - \mathbf{h'}^{\mathcal{P}}_i)$.
Thus, the regression loss $\mathcal{L}_{reg}$ can be represented as:
\begin{equation}
    \mathcal{L}_{reg} = \frac{1}{N} \sum^N_{i=1} \, (1 - {y_i}^\mathsf{T}\hat{y_i}).
\end{equation}

Finally, the overall objective of the point cloud SSRL is to optimize PCExpert with the combination of the above three losses as:
\begin{equation}
    \mathcal{L} = \mathcal{L}_{cm} + \mathcal{L}_{im} + \mathcal{L}_{reg}.
\end{equation}

\section{Experiments}
\label{s:experiment}

\noindent The pre-training of our PCExpert is conducted on the ShapeNet~\cite{chang2015shapenet} dataset, using the methodology described in Section~\ref{s:methods}.
Comprehensive details regarding this pre-training setup are described in Section~\ref{s:experiment:setup}.
We subsequently evaluate the pre-trained model across a variety of 3D point cloud classification benchmarks.
Prior to this evaluation, the model is fine-tuned on each downstream task.
The model's performance on these downstream tasks is reported in Section~\ref{s:experiment:evaluation}.
In Section~\ref{s:experiment:ablation}, we engage in a series of ablation studies exploring various aspects of PCExpert.

\subsection{Pre-training Setup}
\label{s:experiment:setup}

\subsubsection{Dataset}
Following previous studies~\cite{afham2022crosspoint,dong2023autoencoders}, we utilize ShapeNet, a dataset encompassing over $50,000$ CAD models across $55$ categories, as the pre-training dataset for PCExpert.
We employ $40,523$ instances across $13$ categories in ShapeNet to generate point and image data triplets.
For point cloud data, we follow~\cite{zhangpoint} and sample $2,048$ points for each instance, and group them into 160 local patches with group size of 32, whose centroids are sampled with FPS.
As described in Section~\ref{s:methods:training}, the intra-modal contrastive loss $\mathcal{L}_{im}$ and the regression loss $\mathcal{L}_{reg}$ are calculated based on the original point data and its transformed version.
To obtain the transformed point, we rotate the original point about the $y$-axis by a predetermined degree between $[0^{\circ} ,360^{\circ}]$ according to the corresponding image in the triplet (described below).

\begin{figure}[!t] 
    \centering
    \includegraphics[width=.8\linewidth]
    {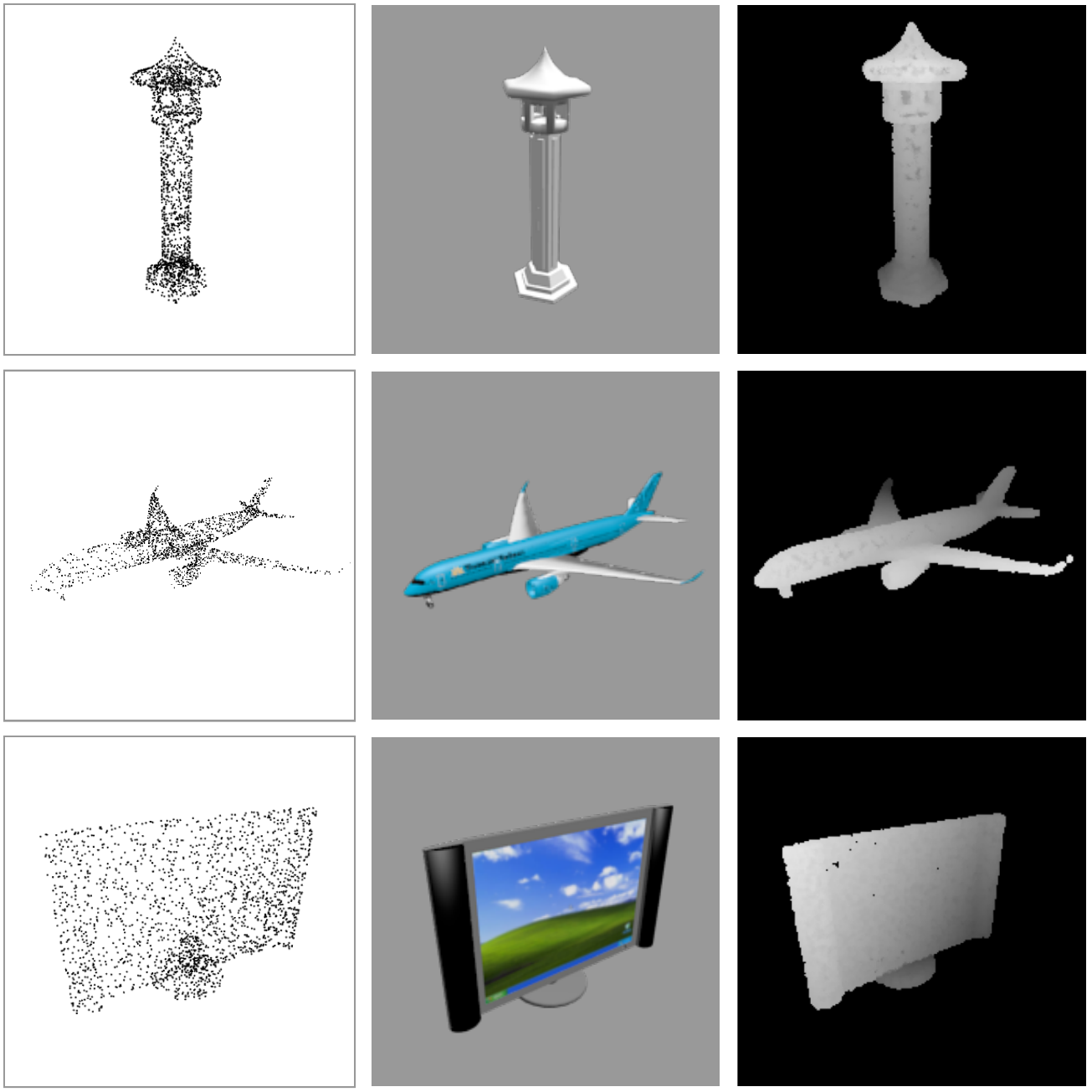}
    \caption{Training samples used in point-image contrastive learning. 
    \textbf{Left}: Point cloud samples. 
    \textbf{Middle}: Images rendered from 3D CAD meshes. 
    \textbf{Right}: Images rendered directly from the original point clouds, with the shape and details well preserved.
    }
    \label{f:sample}
\end{figure}

For image data, we use two types of rendered images.
The first type of images, derived from study~\cite{xu2019disn}, are rendered from the CAD meshes with 36 random views for each mesh.
The second type of images are rendered directly in real time from point data, using the Pytorch3D~\cite{ravi2020pytorch3d} library, with a random rotation angle around the $y$-axis (the yaw angle).
This angle is then recorded and utilized for the corresponding rotation of the point cloud sample.
The dimensions of image data are set to $224 \times 224 \times 3$, with the patch size of $16 \times 16$ and no augmentation applied.
Representative sample images of the two types of rendered images are illustrated in Fig.~\ref{f:sample}.

\subsubsection{Model}
We adopt the image tower of the CLIP model~\cite{radford2021learning} as the base ViT for point and image data, which consists of 12-layer Transformer blocks, with 768 hidden size and 12 attention heads.
The PCExpert module is applied on each Transformer block with a projection dimension of 192.

The model is pre-trained for 300 epochs, with a batch size of 1024.
AdamW~\cite{loshchilov2018decoupled} optimizer is employed with $\beta_1 = 0.9$, $\beta_2 = 0.98$.
The learning rate is initialized to $1e-3$ for the model which uses the mesh-rendered images, and $5e-4$ for the model using the point-rendered images, with both weight decays set to $0.01$.
The training incorporates a linear warmup over the first 400 steps, followed by a cosine decay.

\subsection{Evaluation}\label{s:experiment:evaluation}

\begin{table*}[!t]
\caption{Classification results on ScanObjectNN. 
ModelNet40$-$1k signifies that $1,024$ points are sampled from each sample during the training and test stages. 
$^{\ast}$: Results based on Support Vector Machines (SVMs). 
\texttt{CL}: Methods that are based on contrastive learning are marked with $\surd$.
The overall accuracy (\%) is reported. }
\label{t:sonn}
\begin{center}
\begin{tabular}{ l c c c c c c }
\hline
\multirow{2}{*}{Method} 
&\multirow{2}{*}{\texttt{CL}}
&\multirow{2}{*}{\#Params (M)} 
&\multicolumn{3}{c}{ScanObjectNN} 
&\multirow{2}{*}{ModelNet40$-$1k}\\
\cline{4-6}
 & & &OBJ\textunderscore BG &OBJ\textunderscore ONLY &PB\textunderscore T50\textunderscore RS &\\
\hline
\multicolumn{7}{c}{\textit{Supervised Learning Only}} \\
\hline
PointNet~\cite{qi2017pointnet}          
    &&3.5    &73.3   &79.2   &68.0           &89.2 \\
PointNet++~\cite{qi2017pointnet++}      
    &&1.5    &82.3   &84.3   &77.9           &90.7 \\    
DGCNN~\cite{wang2019dynamic}            
    &&1.8    &82.8   &86.2   &78.1           &92.9 \\
PointCNN~\cite{li2018pointcnn}          
    &&0.6    &86.1   &85.5   &78.5           &92.2 \\
GBNet~\cite{qiu2021geometric}           
    &&8.8    &-      &-      &80.5           &93.8 \\
PointMLP~\cite{marethinking}            
    &&12.6   &-      &-      &85.4$\pm$0.3   &94.1 \\
PointNeXt~\cite{qian2022pointnext}      
    &&1.4    &-      &-      &87.7$\pm$0.4   &93.2 \\
\hline
\multicolumn{7}{c}{\textit{with Self-supervised Representation Learning} (\texttt{FULL})} \\
\hline
MVR~\cite{tran2022self}                 
    &$\surd$&1.8    &84.5$\pm$0.6   &84.3$\pm$0.6   &-      &93.2$\pm$0.1\\
CrossNet~\cite{wu2023self}
    &$\surd$&1.8    &-  &-  &-  &93.4\\
Point-LGMask~\cite{tang2023point}
    &$\surd$&-      &89.8   &89.3   &85.3   &-\\
Transformer~\cite{vaswani2017attention} 
    &&22.1   &83.04          &84.06          &79.11          &91.4 \\
OcCo~\cite{wang2021unsupervised}        
    &&22.1   &84.85          &85.54          &78.79          &92.1 \\
Point-BERT~\cite{yu2022point}           
    &&22.1   &87.43          &88.12          &83.07          &93.2 \\
Point-MAE~\cite{pang2022masked}         
    &&22.1   &90.02          &88.29          &85.18          &93.8 \\
Point-M2AE~\cite{zhangpoint}            
    &&15.3   &91.22          &88.81          &86.43          &94.0 \\
ACT~\cite{dong2023autoencoders} 
    &&22.1   &92.48$\pm$0.59 &91.57$\pm$0.37 &87.88$\pm$0.36 &93.7 \\
\rowcolor{tablehighlight}\textbf{PCExpert (Ours)} 
&$\surd$
&\textbf{6.1} 
&\textbf{92.66$\pm$0.36} 
&91.39$\pm$0.17 
&87.10$\pm$0.20 
&92.7\\
\textit{Improvement} & & &\textcolor{ForestGreen}{($\uparrow 0.18$)} & & &\\
\hline
\multicolumn{7}{c}{\textit{with Self-supervised Representation Learning} (\texttt{LINEAR})} \\
\hline
CrossPoint$^{\ast}$~\cite{afham2022crosspoint}   
    &$\surd$&1.8    &81.7           &-              &-              &91.2 \\
CrossNet$^{\ast}$~\cite{wu2023self}
    &$\surd$&1.8    &83.9   &-  &-  &91.5\\
Point-MAE~\cite{pang2022masked}    
    &&22.1   &82.58$\pm$0.58 &83.52$\pm$0.41 &73.08$\pm$0.30 &91.22$\pm$0.26 \\
ACT~\cite{dong2023autoencoders}    
    &&22.1   &85.20$\pm$0.83 &85.84$\pm$0.15 &76.31$\pm$0.26 &91.36$\pm$0.17 \\
\rowcolor{tablehighlight}\textbf{PCExpert (Ours)} 
&$\surd$
&\textbf{6.1} 
&\textbf{90.02$\pm$0.34} 
&\textbf{89.56$\pm$0.20} 
&\textbf{79.42$\pm$0.10}  
&\textbf{92.22$\pm$0.11}\\
\textit{Improvement} & &
&\textcolor{ForestGreen}{($\uparrow 4.82$)} 
&\textcolor{ForestGreen}{($\uparrow 3.72$)} 
&\textcolor{ForestGreen}{($\uparrow 3.11$)} 
&\textcolor{ForestGreen}{($\uparrow 0.72$)} \\
\hline
\multicolumn{7}{c}{\textit{with Self-supervised Representation Learning} (\texttt{MLP3})} \\
\hline
Point-MAE~\cite{pang2022masked}    
    &&22.1   &84.29$\pm$0.55  &85.24$\pm$0.67  &77.34$\pm$0.12  &92.33$\pm$0.09\\
ACT~\cite{dong2023autoencoders}    
    &&22.1   &87.14$\pm$0.22  &88.90$\pm$0.40  &81.52$\pm$0.19  &92.69$\pm$0.18\\
\rowcolor{tablehighlight}\textbf{PCExpert (Ours)} 
&$\surd$
&\textbf{6.1} 
&\textbf{89.96$\pm$0.43} 
&\textbf{89.76$\pm$0.42} 
&\textbf{82.57$\pm$0.62}   
&\textbf{92.73$\pm$0.12}\\
\textit{Improvement} & & 
&\textcolor{ForestGreen}{($\uparrow 2.82$)}   
&\textcolor{ForestGreen}{($\uparrow 0.86$)}  
&\textcolor{ForestGreen}{($\uparrow 1.05$)} 
&\textcolor{ForestGreen}{($\uparrow 0.04$)} \\
\hline
\end{tabular}
\end{center}
\end{table*}

\subsubsection{Fine-tuning}
\label{s:experiment:evaluation:finetune}

\noindent In this study, fine-tuning is conducted on several widely used classification datasets: ScanObjectNN~\cite{uy2019revisiting} and ModelNet40~\cite{wu20153d}.
ScanObjectNN, a challenging point cloud dataset, comprises $2,880$ objects spanning 15 categories, all generated via scanning real indoor objects.
As is common practice, three variants of this datasets are used in this study, \textit{i.e.}, 
1) OBJ\textunderscore ONLY: the vanilla dataset including only segmented objects; 
2) OBJ\textunderscore BG: a noisier variant including objects with their background elements; and
3) PB\textunderscore T50\textunderscore RS: the most challenging perturbed variant, where each instance is extracted from a bounding box that is randomly shifted up to 50\% of its original size from the ground-truth, in addition to random rotation and scaling. 
ModelNet40 is a synthetic dataset, produced by sampling from 3D CAD models, featuring $12,331$ objects across 40 categories.
Consistent with established practice, we sample $1,024$ points from each instance in the dataset during fine-tuning and test stages, and the results are denoted by ModelNet40$-$1k. 

During the fine-tuning stage, the $[\mathtt{CLS}]$ token from the final output representation is used as the global representation of the sample.
Classification heads are employed to project the representation into the target class space.
We follow the common
practice in~\cite{dong2023autoencoders} and conduct three fine-tuning protocols:
\begin{itemize}
    \item \texttt{FULL}: All the parameters of the PCExpert module and the classification head are updated in fine-tuning, while other parameters in ViT are kept frozen. The classification head is a three-layer non-linear MLP. 
    \item \texttt{LINEAR}: Only the parameters of the classification head are updated during fine-tuning, which is a single-layer MLP.
    \item \texttt{MLP3}: Only the parameters of the classification head are updated during fine-tuning. This is also a three-layer non-linear MLP (\textit{i.e.}, the same as that in the protocol \texttt{FULL}).
\end{itemize}

\subsubsection{3D Object Classification}\label{s:experiment:evaluation:sonn}

\noindent The classification results of ScanObjectNN and ModelNet40 are presented in Table~\ref{t:sonn}.

Firstly, it can be observed that our PCExpert outperforms the existing state-of-the-art (SOTA) SSRL methods across all benchmarks in \texttt{LINEAR} and \texttt{MLP3} protocols, especially for the challenging real-world dataset ScanObjectNN, where it achieves the highest accuracy improvement of $+4.8\%$ in \texttt{LINEAR}.
Because the majority of model parameters are not updated during \texttt{LINEAR} fine-tuning, the performance on this benchmark heavily relies on the model's generalizability and understanding of the underlying point cloud semantics.
This attests to PCExpert's exceptional representation capabilities.
With a much smaller model size, PCExpert can still achieve performance comparable to other models under benchmarks using the \texttt{FULL} protocol.

Secondly, in comparison to the studies based on point-image contrastive learning (\textit{e.g.}, CrossPoint~\cite{afham2022crosspoint} and MVR~\cite{tran2022self}), PCExpert significantly outperforms the existing methods, with average improvements of $+4.9\%$ and $+8.3\%$ in the \texttt{FULL} and \texttt{LINEAR} protocols, respectively.

Thirdly, it is noted that,  PCExpert exhibits minimal inductive bias concerning 3D understanding (through the patch embedding module), and yet its performance still surpasses models that are characterized by this specific inductive bias, such as Point-M2AE~\cite{zhangpoint}, across the majority of the benchmarks.

Furthermore, our findings show that PCExpert demonstrates better performance on the real-world dataset ScanObjectNN compared to the synthetic ModelNet40. 
We postulate this superiority is likely a result of the effective knowledge transfer from CLIP, leveraging the vast quantity of image-based training data.

\begin{table*}[!t]
\caption{Few-shot classification results on ModelNet40. The overall accuracy (\%) is reported.}
\label{t:mn40fs}
\begin{center}
\begin{tabular}{ l c c c c }
\hline
\multirow{2}{*}{Method} &\multicolumn{2}{c}{5-way} &\multicolumn{2}{c}{10-way} \\
\cmidrule(r{2pt}){2-3} \cmidrule(l{2pt}){4-5}
&10-shot &20-shot &10-shot &20-shot \\
\hline
\multicolumn{5}{c}{\textit{with Self-supervised Representation Learning} (\texttt{FULL})} \\
\hline
CrossPoint~\cite{afham2022crosspoint}        
    &92.5 $\pm$ 3.0 &94.9 $\pm$ 2.1 &83.6 $\pm$ 5.3 &87.9 $\pm$ 4.2 \\
Point-LGMask~\cite{tang2023point}
    &97.4 $\pm$ 2.0 &98.1 $\pm$ 1.4 &92.6 $\pm$ 4.3 &95.1 $\pm$ 3.4 \\
Transformer~\cite{vaswani2017attention}     
    &87.8 $\pm$ 5.2 &93.3 $\pm$ 4.3 &84.6 $\pm$ 5.5 &89.4 $\pm$ 6.3 \\
OcCo~\cite{wang2021unsupervised}            
    &94.0 $\pm$ 3.6 &95.9 $\pm$ 2.3 &89.4 $\pm$ 5.1 &92.4 $\pm$ 4.6 \\
Point-BERT~\cite{yu2022point}      
    &94.6 $\pm$ 3.1 &96.3 $\pm$ 2.7 &91.0 $\pm$ 5.4 &92.7 $\pm$ 5.1 \\
Point-MAE~\cite{pang2022masked}       
    &96.3 $\pm$ 2.5 &97.8 $\pm$ 1.8 &92.6 $\pm$ 4.1 &95.0 $\pm$ 3.0 \\
Point-M2AE~\cite{zhangpoint}       
    &96.8 $\pm$ 1.8 &98.3 $\pm$ 1.4 &92.3 $\pm$ 4.5 &95.0 $\pm$ 3.0 \\
ACT~\cite{dong2023autoencoders}             
    &96.8 $\pm$ 2.3 &98.0 $\pm$ 1.4 &93.3 $\pm$ 4.0 &95.6 $\pm$ 2.8 \\
\rowcolor{tablehighlight}\textbf{PCExpert (Ours)} 
&\textbf{98.0 $\pm$ 1.8}
&\textbf{98.8 $\pm$ 0.9}
&\textbf{93.8 $\pm$ 4.4}
&\textbf{96.2 $\pm$ 3.0} \\
\textit{Improvement}  
&\textcolor{ForestGreen}{($\uparrow 0.6$)}   
&\textcolor{ForestGreen}{($\uparrow 0.7$)}  
&\textcolor{ForestGreen}{($\uparrow 0.5$)} 
&\textcolor{ForestGreen}{($\uparrow 0.6$)}  \\
\hline
\multicolumn{5}{c}{\textit{with Self-supervised Representation Learning} (\texttt{LINEAR})} \\
\hline
Point-MAE~\cite{pang2022masked} 
    &91.1 $\pm$ 5.6 &91.7 $\pm$ 4.0 &83.5 $\pm$ 6.1 &89.7 $\pm$ 4.1 \\
ACT~\cite{dong2023autoencoders} 
    &91.8 $\pm$ 4.7 &93.1 $\pm$ 4.2 &84.5 $\pm$ 6.4 &90.7 $\pm$ 4.3 \\
\rowcolor{tablehighlight}\textbf{PCExpert (Ours)} 
&\textbf{97.2 $\pm$ 1.9}
&\textbf{97.7 $\pm$ 1.4}
&\textbf{92.9 $\pm$ 4.2}
&\textbf{94.8 $\pm$ 3.4} \\
\textit{Improvement}  
&\textcolor{ForestGreen}{($\uparrow 5.4$)}   
&\textcolor{ForestGreen}{($\uparrow 4.6$)}  
&\textcolor{ForestGreen}{($\uparrow 8.4$)} 
&\textcolor{ForestGreen}{($\uparrow 4.1$)}  \\
\hline
\multicolumn{5}{c}{\textit{with Self-supervised Representation Learning} (\texttt{MLP3})} \\
\hline
Point-MAE~\cite{pang2022masked}
    &95.0 $\pm$ 2.8 &96.7 $\pm$ 2.4 &90.6 $\pm$ 4.7 &93.8 $\pm$ 5.0 \\
ACT~\cite{dong2023autoencoders} 
    &95.9 $\pm$ 2.2 &97.7 $\pm$ 1.8 &92.4 $\pm$ 5.0 &94.7 $\pm$ 3.9 \\
\rowcolor{tablehighlight}\textbf{PCExpert (Ours)} 
&\textbf{97.0 $\pm$ 2.6}
&\textbf{98.5 $\pm$ 1.0}
&\textbf{92.8 $\pm$ 3.8}
&\textbf{95.5 $\pm$ 2.9} \\
\textit{Improvement}  
&\textcolor{ForestGreen}{($\uparrow 1.1$)}   
&\textcolor{ForestGreen}{($\uparrow 0.8$)}  
&\textcolor{ForestGreen}{($\uparrow 0.4$)} 
&\textcolor{ForestGreen}{($\uparrow 0.8$)}  \\
\hline
\end{tabular}
\end{center}
\end{table*}

\subsubsection{Few-shot Point Cloud Classification} \label{s:experiment:evaluation:mn40}

\noindent The results of our few-shot 3D object classification experiments are summarized in Table~\ref{t:mn40fs}. 

Several key findings are as follows:
First, our PCExpert consistently outperforms the existing methods across all experiments. 
Specifically, significant performance gains 
of +4\% to +8\% are noted in \texttt{LINEAR}, and +4\% to +10\% improvement compared with point-image contrastive methods (\textit{e.g.}, CrossPoint~\cite{afham2022crosspoint}).

Secondly, it can be observed that our PCExpert's performance under the \texttt{LINEAR} protocol begins to approach that under the \texttt{FULL} protocol. 
This observation implies that the model is already robustly generalizable after pre-training, and requires only linear projections for effective application. 
This suggests a decreasing need for extensive fine-tuning of the whole model parameters on specific tasks, a process which typically demands significant time and resources.
Interestingly, we observe that as the number of training samples decreases, our model's superiority over the existing SOTA becomes more apparent. 
For instance, under \texttt{LINEAR}, the performance improvement in 10-shot settings is consistently higher than that in 20-shot settings.
This phenomenon indicates that PCExpert can extract and use meaningful features more effectively, while remaining robust to overfitting.
This further substantiates the representation capability and generalizability of PCExpert.

These results collectively indicate that our method provides a robust and effective 
approach for few-shot classification, demonstrating notable improvements over existing techniques, even with reduced trainable parameters.

\subsection{Ablation Studies}
\label{s:experiment:ablation}

\subsubsection{Parameter Sharing}
Based on our novel idea of reinterpreting point clouds as specialized images, we propose that extensive parameter sharing with image encoders can be beneficial for point cloud understanding.
To validate this,
we compare the performance 
between the proposed PCExpert and separate point encoders (\textit{e.g.}, Transformer and DGCNN~\cite{wang2019dynamic}), as shown in Table~\ref{t:dgcnn}.
The comparisons are based on the same pre-training dataset and objectives, with the only difference being whether ViT participates in point data encoding. 
For the separated encoders, the losses are calculated based on the point output representations from the Transformer or DGCNN, and the image output representations from ViT, where ViT does not access or process point data.

It can be seen from the results that PCExpert exhibits better performance in both evaluations, despite the inductive bias of the separate encoder (\textit{e.g.}, DGCNN~\cite{wang2019dynamic}).
This outcome strongly suggests the crucial contribution of parameter sharing in image knowledge transfer, which thereby enhances the model's representation generalizability.
Notably, given that Transformer architectures possess advantages in: 1) end-to-end learning on data from heterogeneous modality, and 2) the scalability with increased computational resources, our approach in this study presents a promising direction for future multi-modal studies and applications on point clouds.

\begin{table}[!t]
\caption{Ablation on network architecture. 
SO-BG: the OBJ\textunderscore BG split of the ScanObjectNN dataset.
MN-1k: the 1k-sampling setting of the ModelNet40 dataset.
The overall accuracy (\%) under the \texttt{LINEAR} protocol is reported.}
\label{t:dgcnn}
\begin{center}
\begin{tabular}{ l c c c }
\hline
Method          &Parameter Sharing &SO-BG              &MN-1k\\
\hline
Transformer     &$\times$       &86.19$\pm$0.26     &89.42$\pm$0.12\\
DGCNN           &$\times$       &88.73$\pm$0.43     &89.44$\pm$0.26\\
PCExpert        &$\surd$        &90.02$\pm$0.34     &92.22$\pm$0.11\\ 
\hline  
\end{tabular}
\end{center}
\end{table}

\begin{table}[!t]
\caption{Comparison of PCExpert performance pre-trained with mesh rendered (PCExpert-M) and point cloud rendered images (PCExpert-P). 
SO-BG, SO-OBJ, and SO-PB: the OBJ\textunderscore BG, the OBJ\textunderscore ONLY, and the PB\textunderscore T50\textunderscore RS variants of the ScanObjectNN dataset, respectively.
MN-1k: the 1k-sampling setting of the ModelNet40 dataset.
MN-$i$w$j$s: the $i$-way $j$-shot few-shot setting of the ModelNet40 dataset.
The overall accuracy (\%) is reported.}
\label{t:pc_rendered}
\begin{center}
\begin{tabular}{ l l c c }
\hline
Protocol    &Benchmark                 &PCExpert-M         &PCExpert-P \\
\hline
\multirow{8}{*}{\texttt{FULL}}
    &SO-BG          &91.91                      &\textbf{92.66}\\
    &SO-OBJ         &91.22                      &\textbf{91.39}\\
    &SO-PB          &87.09                      &\textbf{87.10}\\
    &MN-1k          &92.50                      &\textbf{92.67}\\ 
    &MN-5w10s       &\textbf{98.0$\pm$1.8}      &96.5 $\pm$ 2.7\\
    &MN-5w20s       &\textbf{98.8$\pm$0.9}      &98.0 $\pm$ 1.5\\
    &MN-10w10s      &\textbf{93.8$\pm$4.4}      &93.2 $\pm$ 4.6\\
    &MN-10w20s      &\textbf{96.2$\pm$3.0}      &95.6 $\pm$ 3.2\\
\hline
\multirow{8}{*}{\texttt{LINEAR}}
    &SO-BG          &\textbf{90.02}             &88.30\\
    &SO-OBJ         &\textbf{89.56}             &87.09\\
    &SO-PB          &\textbf{79.42}             &77.03\\
    &MN-1k          &\textbf{92.22}             &91.33\\
    &MN-5w10s       &\textbf{97.2$\pm$1.9}      &97.0 $\pm$ 2.8\\
    &MN-5w20s       &\textbf{97.7$\pm$1.4}      &97.0 $\pm$ 2.2\\
    &MN-10w10s      &\textbf{92.9$\pm$4.2}      &90.8 $\pm$ 5.2\\
    &MN-10w20s      &\textbf{94.8$\pm$3.4}      &93.1 $\pm$ 4.3\\
\hline
\multirow{8}{*}{\texttt{MLP3}}
    &SO-BG          &89.96                      &\textbf{90.53}\\
    &SO-OBJ         &89.76                      &\textbf{89.85}\\
    &SO-PB          &\textbf{82.57}             &81.96\\
    &MN-1k          &\textbf{92.73}             &92.34\\
    &MN-5w10s       &\textbf{97.0$\pm$2.6}      &96.9 $\pm$ 2.5\\
    &MN-5w20s       &\textbf{98.5$\pm$1.0}      &97.8 $\pm$ 1.5\\
    &MN-10w10s      &\textbf{92.8$\pm$3.8}      &91.2 $\pm$ 4.8\\
    &MN-10w20s      &\textbf{95.5$\pm$2.9}      &94.2 $\pm$ 4.1\\
\hline  
\end{tabular}
\end{center}
\end{table}

\subsubsection{Point Cloud Rendered Images}
Table~\ref{t:pc_rendered} shows the performance of PCExpert pre-trained using point cloud rendered images (dubbed ``PCExpert-P'').

Our experimental results reveal that PCExpert, pre-trained using mesh-rendered data (dubbed ``PCExpert-M''), consistently exhibits superior performance in the majority of benchmarks, particularly in the few-shot experiments and those under the \texttt{LINEAR} protocol.
This suggests that the exploitation of mesh-rendered images, which more closely resembles real-world photos, promotes more effective learning of robust representation capabilities.
Nonetheless, it is important to highlight that PCExpert-P, in some experiments under the \texttt{FULL} protocol, 
demonstrates superior performance to PCExpert-M.
This can be attributed to the pre-training that is solely based on the single modality of point clouds, which makes it easier for the model to optimize in tasks related to point clouds.

Moreover, while being translated into a suitable form for image encoders, the point cloud-rendered images have preserved crucial semantic characteristics of the original point clouds, as shown in Fig.~\ref{f:sample}.
As a result, the performance differential between the two models is marginal, and PCExpert-P also surpasses existing state-of-the-art benchmarks in numerous experiments. 
Considering that utilizing point cloud-rendered images can significantly reduce dataset creation costs and difficulties, the minor performance deficiencies can be compensated by leveraging larger quantities of data in the absence of mesh-rendered images.

\begin{table}[!t]
\caption{Ablation on pre-training objectives. 
Overall accuracy (\%) on the ScanObjectNN OBJ\textunderscore BG (SO-BG) benchmark under the \texttt{LINEAR} protocol are reported.}
\label{t:objectives}
\begin{center}
\begin{tabular}{ c c c c }
\hline
$\mathcal{L}_{cm}$    
&$\mathcal{L}_{im}$   
&$\mathcal{L}_{reg}$
&\multirow{2}{*}{SO-BG} \\
(cross-modal)
&(intra-modal) 
&(regression)
&\\
\hline
$\surd$     &$\times$   &$\times$   &84.71$\pm$0.18\\
$\surd$     &$\surd$    &$\times$   &82.96$\pm$0.12\\
$\surd$     &$\times$   &$\surd$    &85.48$\pm$0.17\\
$\surd$     &$\surd$    &$\surd$    &90.02$\pm$0.34\\
\hline  
\end{tabular}
\end{center}
\end{table}

\begin{figure}[!t] 
    \centering
    \includegraphics[width=.9\linewidth]
    {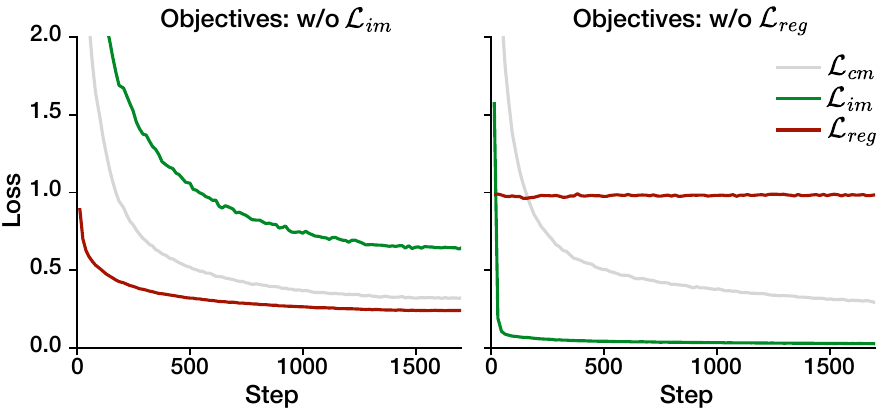}
    \caption{\textbf{left}: Gradient calculation is based on  $\mathcal{L}_{cm}$ and $\mathcal{L}_{reg}$, excluding $\mathcal{L}_{im}$. 
    Optimizing for $\mathcal{L}_{reg}$ (the red curve) concurrently results in a reduction of $\mathcal{L}_{im}$ (green).
    \textbf{right}: Gradient calculation is based on $\mathcal{L}_{cm}$ and $\mathcal{L}_{im}$, excluding $\mathcal{L}_{reg}$. 
    Optimizing for $\mathcal{L}_{im}$ has no effect on $\mathcal{L}_{reg}$.}
    \label{f:regression}
\end{figure}

\subsubsection{Pre-training Objectives}

We conduct an ablation study to assess the significance of different pre-training objectives. 
The results are summarized in Table~\ref{t:objectives}.

Our analysis reveals that the inclusion of the regression loss $\mathcal{L}_{reg}$ yields better performance in both scenarios of $\mathcal{L}_{cm}+\mathcal{L}_{reg}$ and $\mathcal{L}_{cm}+\mathcal{L}_{im}+\mathcal{L}_{reg}$.
These results provide evidence that the parameter estimation task contributes to enhancing the representation capability of the model.

However, we have observed a notable decrease in performance when incorporating $\mathcal{L}_{im}$ with $\mathcal{L}_{cm}$, a finding that stands in contrast to that reported in study~\cite{afham2022crosspoint}.
The discrepancy can likely be attributed to issues stemming from loss balancing within our architecture, as the intra-modal contrastive task might be more challenging for the ViT model employed in our study compared to the 3D-specific encoder (\textit{i.e.}, DGCNN~\cite{wang2019dynamic}) used in study~\cite{afham2022crosspoint}.
This difficulty may cause the model to focus excessively on minimizing $\mathcal{L}_{im}$, thereby neglecting $\mathcal{L}_{cm}$ and compromising the generalizability of the overall representation.

Interestingly, when $\mathcal{L}_{reg}$ is introduced into the mix, it appears to interactively reduce the difficulty of optimizing for $\mathcal{L}_{im}$.
As illustrated in Fig.~\ref{f:regression}, when optimizing for $\mathcal{L}_{reg}$, there is an inherent reduction in $\mathcal{L}_{im}$ (the left plot in Fig.~\ref{f:regression}), even though the latter loss is intentionally excluded from the gradient calculation.
However, the reverse relationship is not true (the right plot in Fig.~\ref{f:regression}).
This discovery suggests that the characteristics of point cloud learned through $\mathcal{L}_{reg}$ contribute to the objective of $\mathcal{L}_{im}$, establishing a beneficial synergy between the two losses.
This interplay results in less optimization difficulty and, consequently, optimal model performance.

\section{Conclusion}
\label{s:conclusion}

\noindent In this study, we have proposed PCExpert, 
a novel architecture for point cloud self-supervised representation learning.
By reconsidering point cloud data as ``specialized images'', we have opened up a new pathway of utilizing large-scale image knowledge to enhance point cloud understanding, \textit{i.e.}, through extensive parameter sharing with a pre-trained ViT.
The sharing is enabled via the multi-way Transformer blocks which share the Multi-head Self-Attention (MSA) modules of the ViT, while providing separate feed forward network (FFN) modules dedicated to the learning of point knowledge.

This design facilitates an in-depth knowledge transfer from images to point clouds, and thereby significantly augments point cloud representation capabilities.
Together with the introduction of a novel pretext task that leverages the ``regression loss'', PCExpert has demonstrated remarkable performance across multiple benchmarks, especially in experiments under the \texttt{LINEAR} fine-tuning protocol and in few-shot scenarios.
PCExpert's performance serves as a solid validation for our proposition of reconsidering point clouds as images. 
This standpoint is also reflected in our novel approach of generating contrastive images directly from point cloud rendering. 
This opens up new possibilities for augmenting the size of point cloud datasets for contrastive learning.

By these means, this study indicates a promising direction for future studies on point-image contrastive representation learning, through parameter sharing, coupled with larger datasets and the scalability of Transformers.

\bibliographystyle{plain}
\bibliography{ref.bib}

\end{document}